\documentclass[10pt,twocolumn,letterpaper]{article}

\usepackage{wacv}              

\definecolor{wacvblue}{rgb}{0.21,0.49,0.74}
\usepackage[pagebackref,breaklinks,colorlinks,allcolors=wacvblue]{hyperref}

\def\wacvPaperID{*****} 
\def\confName{WACV}
\def\confYear{2026}

\title{Latent Inter-Frame Pruning: A Training-Free Method Bridging Traditional Video Compression and Modern Diffusion Transformers for Efficient Generation}

\author{Dennis Menn\\
The University of Texas at Austin \thanks{Work partially done during an internship at Futurewei Technologies, Inc.} \\
dennismenn@utexas.edu 
\and
Chih-Hsien Chou\\
Futurewei Technologies, Inc.\\
cchou@futurewei.com
}

\begin{document}
\maketitle
\begin{abstract}
Video generation, while capable of generating realistic videos, is computationally expensive and slow, prohibiting real-time applications. In this paper, we observe that video latents encoded via an autoencoder under the Latent Diffusion Model (LDM) framework contain redundancy along the temporal axis. Analogous to how traditional video compression algorithms avoid transmitting redundant frame data, we propose the \textbf{Latent Inter-frame Pruning} framework to prune (skip the re-computation of) duplicated latent patches, thereby reducing computational burden and increasing throughput. However, direct pruning results in visual artifacts due to the discrepancy between full-sequence training and pruned inference. To resolve these artifacts, we propose an \textbf{Attention Recovery} mechanism to bridge the train-inference gap. With our proposed method, we increase video editing throughput by 1.44$\times$, achieving 12.44 FPS on an NVIDIA RTX 6000 while maintaining video quality. We hope our work inspires further research into integrating traditional video compression methods with modern video generation pipelines. This work is a preliminary work on Training-free Latent Inter-Frame Pruning with Attention Recovery.

\end{abstract}
    
\section{Introduction}
\label{sec:intro}
Videos are composed of consecutive image frames that inherently contain redundancy across both temporal and spatial dimensions. Traditional video compression technologies exploit this property by identifying and compressing duplicated information \cite{MPEG1991}, thereby reducing the volume of data required for representation. Consequently, this lowers the computational budget and bandwidth for data transmission by avoiding processing redundant information.

In this paper, we start by observing that the video latent, compressed via an autoencoder \cite{connor2021vae}, still contains redundancy along the temporal axis. We then adapt the concept of inter-frame compression from video codec \cite{MPEG1991} to generative AI for video editing within the modern Latent Diffusion Models (LDM) framework \cite{rombach2021ldm}. 

Our method performs end-to-end pruning along the temporal dimension. Specifically, instead of redundantly re-generating unchanged latent patches, determined by comparing patches to their spatially corresponding patches from the previous frame, we directly reuse the generated results. This, in turn, lowers the computations needed for video editing tasks and hence increases the throughput for video editing. Additionally, our method is training-free and can be seamlessly integrated into Diffusion Transformer frameworks with causal attention \cite{yin2025causvid, huang2025selfforcing}. 

Several previous studies focused on token pruning within the Transformer architecture \cite{choudhury2024rlt, bolya2022tome, wu2025importancetome}. However, the majority of existing research focuses on the video understanding tasks \cite{choudhury2024rlt, bolya2022tome}. The fundamental distinction between pruning for understanding versus generation lies in the tolerance for information loss and the need to recover the pruned tokens.

\begin{itemize}
	\item \textbf{Understanding Tasks:} In these sparse prediction tasks, tolerance for information loss is high. The objective is to extract high-level semantics—such as classification or movement—meaning the model does not need to preserve every visual detail to succeed. Also, there is no need to recover the pruned tokens to match the output token count with the input token count.
	\item \textbf{Generation Tasks:} Conversely, video generation/editing requires a dense correspondence, where every input patch in latent space maps to a specific spatial region in the output. Therefore, every detail of information will influence the final video quality. This is particularly obvious in sensitive regions like human eyes; a loss of information in these regions may appear numerically small but can lead to significant perceptual degradation. Furthermore, due to the need for decoding, it is necessary to reconstruct pruned tokens to match the output token number to the input token number. 
\end{itemize}. 

While some research explores token pruning for generation tasks, their methods mainly rely on merging and unmerging tokens within specific attention blocks to mitigate the pruning errors \cite{bolya2023tomesd, wu2025importancetome}. In contrast, our research is based on the observation that latent space contains temporal redundancy. Inspired by traditional video codec compression techniques, we prune input tokens in an end-to-end way. This allows allocating less computation to tokens with less information, enabling high pruning ratios while maintaining generation quality. 

There are several challenges in adopting the proposed Inter-Frame Pruning pipeline for generation tasks. First, the entire pruning pipeline must operate within the LDM architecture, posing constraints on the number of input and output tokens required for encoding and decoding. Second, we must mitigate the train-inference discrepancy caused by pruning. Since the model is trained on full-sequence tokens containing complete latent information, directly pruning the latent patch disrupts this distribution, resulting in visual artifacts. To bridge this gap, we propose \textbf{Attention Recovery} mechanism. Third, we must prevent \textbf{information loss} caused by pruning. Since video generation is a dense prediction task where each token corresponds to a specific spatial region, erroneous pruning can significantly degrade visual quality. To address this, we design a robust algorithm that targets only redundant tokens and effectively reintegrates pruned information to maintain fidelity.

Our contributions are summarized as follows:
\begin{enumerate}
	\item We observe that the latent space encoded by VAE contains redundancy, allowing us to design a framework that adapts traditional inter-frame compression algorithms to modern video generation (editing) methods.
	\item We introduce an Attention Recovery mechanism to mitigate the train-inference discrepancy that leads to artifacts, enabling the direct, training-free application of our method without compromising video quality.
	\item We demonstrate the effectiveness of our method on Self-Forcing model \cite{huang2025selfforcing}, achieving a 1.44$\times$ increase in throughput while maintaining editing quality evaluated across 51 video-prompt pairs from the DAVIS dataset~\cite{davis2017dataset}.
\end{enumerate}
\section{Methods}
\label{sec:formatting}
\subsection{Preliminary}
In this paper, we edit videos under the LDM framework with a Transformer backbone \cite{rombach2021ldm}. Given a source video $x$, we first use an encoder $\mathcal{E}$ to map the video into a compressed latent space $z = \mathcal{E}(x)$, thereby reducing computational complexity. These latents are subsequently tokenized and incorporated with Gaussian noise. During inference, the diffusion transformer $\epsilon_\theta$ will denoise the tokens conditioned on both textual prompt and the original video's structural layout. The denoising process will iterate multiple times. Finally, the clean tokens are projected back to the pixel space via the decoder $\mathcal{D}$. 
\begin{figure*}[htp]
  \centering
  \includegraphics[width=0.9\textwidth]{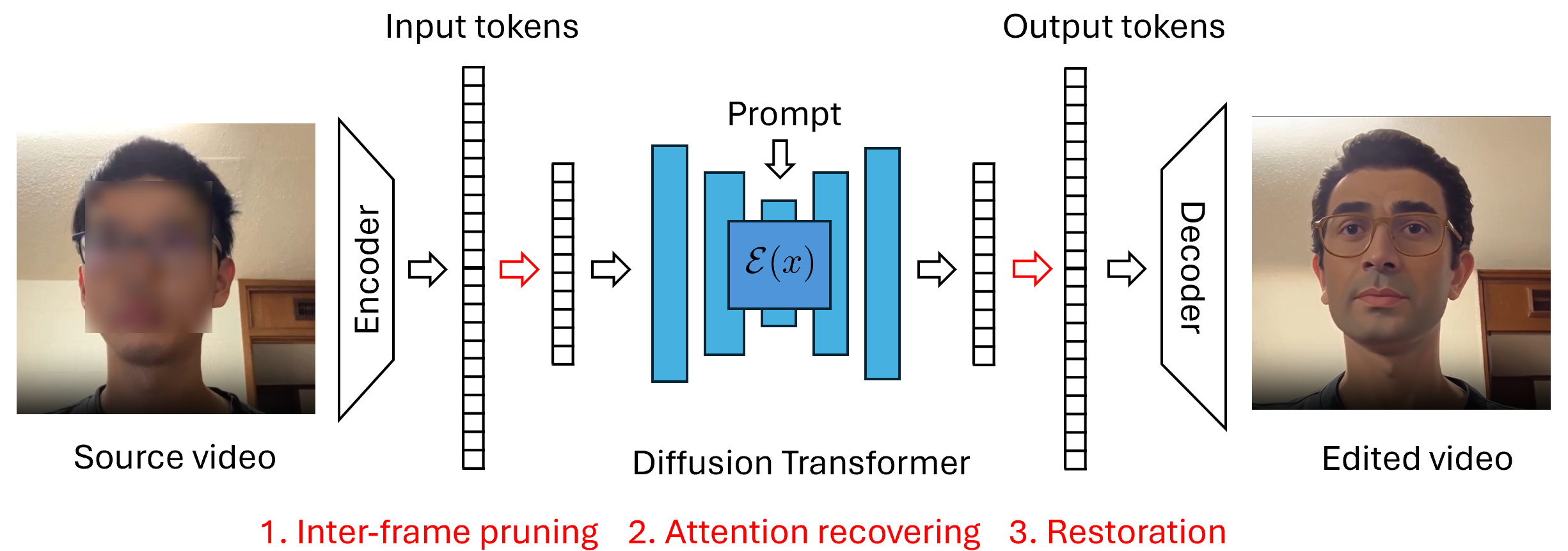}
  \caption{Framework overview: The proposed framework consists of three stages.}
  \label{fig:framework}
\end{figure*}

\subsection{Proposed Framework Overview}
In Figure \ref{fig:framework}, we demonstrate the proposed pruning framework operating in three stages to accelerate the video editing task. First, we apply \textbf{Inter-Frame Pruning} to remove redundant patches in the latent space prior to the tokenization step. Note that each patch will later be transformed into input tokens for the Diffusion Transformer (DiT).  Given the $O(N^2)$ computational complexity of the transformer, where $N$ denotes the number of input tokens, this reduction significantly reduces computational costs. However, directly removing tokens disrupts the input distribution, because when training, the input tokens always contain complete (unpruned) latents information. This difference, caused by pruning, will lead to changes in self-attention values, resulting visual artifacts. To resolve this, we introduce \textbf{Attention Recovery} mechanism, a mathematical approximation that matches attention scores from pruned version with the original (unpruned) calculations. This step incurs negligible overhead and ensures visual quality during the denoising process. Finally, the \textbf{Restoration} step up-samples the tokens number for decoding purpose and can then map latents back to the pixel space.


\subsection{Latent Inter-Frame Pruning}
Natural video exhibits significant temporal redundancy. Inspired by the Inter-Frame compression in video codecs and Run-Length Tokenization (RLT)~\cite{MPEG1991, choudhury2024rlt}, we extend these concepts to the latent space of generative models to bypass the re-generation of redundant tokens. 



Our contribution lies in demonstrating that the latent space of diffusion models maintains the spatiotemporal locality of pixel space, enabling the direct extension of video codec compression algorithms to bypass redundant token regeneration.

Latent Inter-Frame Pruning identifies patches with similar contents by comparing temporally consecutive patches with same spatial location that aligns with the model's tokenization grid. Formally, let $P_{t}^{(x,y)}$ and $P_{t+1}^{(x,y)}$ denote latent patches at spatial coordinates $(x, y)$ across adjacent frames. We define a similarity criterion based on the $L_1$ distance relative to a threshold $\theta$:
\begin{equation}
    \|P_{t}^{(x,y)} - P_{t+1}^{(x,y)}\|_1 < \theta
    \label{eq:pruning_criterion}
\end{equation}
When this condition holds, $P_{t+1}^{(x,y)}$ it is deemed redundant and pruned. 


\subsection{Attention Recovery}
Directly applying Latent Inter-Frame Pruning results in visual artifacts in the generated video. This is because the model is trained exclusively on full-length sequences. Pruning removes a subset of the input tokens, which produces a train-inference discrepancy that degrades the quality of the generation. To address this, we formulate an Attention Recovery mechanism: an analytical approximation that calibrates the pruned output to match full-length inference.

To address the train-inference discrepancy, we first identify which DiT layers are affected by token pruning. The key factor is the \textit{inter-token dependency}: determining which layers compute the output of a target token $x_i'$ based on the presence of other tokens. Mathematically, inter-token dependency can be formulated as:
\begin{equation}
    x_i' = f(x_0, x_1, \dots, x_N)
\end{equation}
In the Transformer block, only the \textbf{Self-Attention} layer satisfies this condition, as it aggregates information from the entire sequence (or a causal window) to update each token. In contrast, Feed-Forward Networks (FFN) and Cross-Attention layers operate point-wise with respect to the input video tokens -- that is, the output $x_i'$ is a function of the token $x_i$ itself (i.e., $x_i' = f(x_i)$). Consequently, the reason why pruning tokens results in visual artifacts originates from self-attention layers.


\begin{figure*}[htp]
	\centering
	\includegraphics[width=0.9\textwidth]{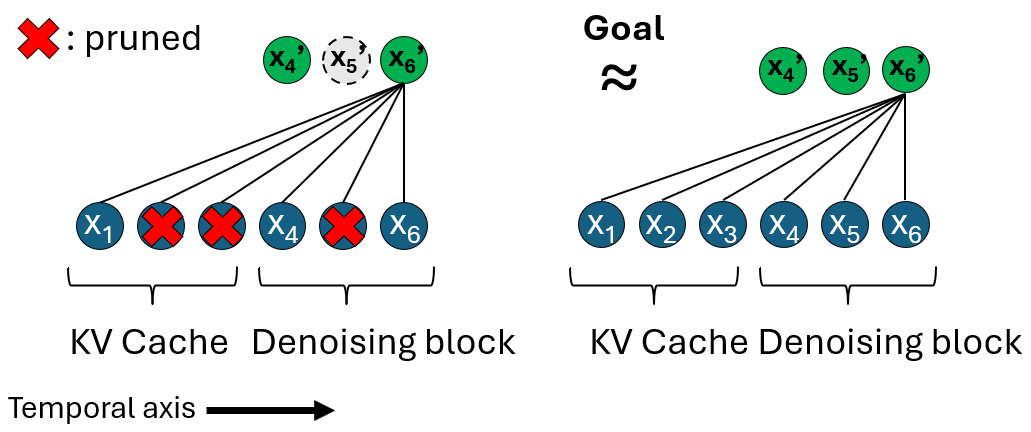}
	\caption{Illustration on approximation of the pruned tokens with full length token sequence. }
	\label{fig:math_goal}
\end{figure*}

In Figure~\ref{fig:math_goal}, we illustrate our objective: to approximate the full-sequence self-attention values using a set of pruned tokens. We conducted our experiments using the causal attention backbone~\cite{yin2025causvid,huang2025selfforcing}. Let $x_i$ denote tokens corresponding to the same spatial location across consecutive temporal frames, such that $x_i$ precedes $x_{i+1}$. In this setup, $\{x_1, x_2, x_3\}$ represents the history stored in the KV cache, while $\{x_4, x_5, x_6\}$ represents the current tokens in the denoising block. Note that the major speed up brought by the Latent Inter-Frame Pruning comes from fewer tokens needed to be generated, throughout all layers in DiT. We will elaborate on this more later.



Mathematically, our goal is to approximate the self-attentions' full-sequence output using only the pruned tokens. We use Figure \ref{fig:math_goal} as an example, where tokens $x_2, x_3$ and $x_5$ are pruned. Our goal is to approximate the following equation:
\begin{equation}
\vspace{2mm}
\hspace{-2mm}
\frac{\sum_{i=1}^6 e^{q^T k_i}V_i}{\sum_{i=1}^6 e^{q^T k_i}} \approx \frac{g(e^{q^T f(k_1, n)})V_1 + g(e^{q^T f(k_4, m)})V_4 + e^{q^T k_6}V_6}{g(e^{q^T f(k_1, n)}) + g(e^{q^T f(k_4, m)}) + e^{q^T k_6}}
\label{eqn:goal}
\end{equation}

where $n=3$ and $m=2$ represent the number of duplicated tokens covered by $x_1$ and $x_4$, respectively. Since the pruned tokens no longer exist in memory, we must use the remaining tokens $\{x_1, x_4, x_6\}$ to recover the original attention calculation. To ensure negligible overhead and compatibility with FlashAttention, we impose a constraint: modifications can only be applied to the Key vectors via a function $f(\cdot)$ or to the attention scores $g(\cdot)$. 

Given that we prune $\{x_2, x_3\}$ and $\{x_5\}$ the Latent Inter-Frame Pruning algorithm implies that the underlying patches are spatially similar, i.e., $P_1 \approx P_2 \approx P_3$ and $P_4 \approx P_5$. Consequently, the corresponding token representations are also similar: $x_1 \approx x_2 \approx x_3$ and $x_4 \approx x_5$.

Note that although positional embedding would differentiate $x_1, x_2, x_3$, our model uses \textbf{Rotary Positional Embeddings (RoPE)}. RoPE largely maintains the property of $x_1 \approx x_2 \approx x_3$ by incorporating positional information that only impacts the attention score calculated from the relevant distance between $q$ and $k$, while leaving $v$ unchanged. 
As a result, we can simplify Eqn. \ref{eqn:goal} by substitution pruned $x_i$ are similar and apply RoPE to $k's$:
\begin{equation}
\label{eqn:goal2}
\resizebox{1.0\columnwidth}{!}{$
\begin{aligned}
   &\frac{\sum_{i=1}^6 e^{q^T k_i}V_i}{\sum_{i=1}^6 e^{q^T k_i}} \approx \\ 
   &\frac{(e^{q^T k_1}+ e^{q^Te^{\theta j} k_1}+ e^{q^T e^{2\theta j} k_1})V_1+ (e^{q^T k_4}+e^{q^Te^{\theta j} k_4})V_4 + e^{q^T k_6}V_6}{e^{q^T k_1} + e^{q^Te^{\theta j} k_1} + e^{q^Te^{2\theta j} k_1} + e^{q^T k_4} + e^{q^Te^{\theta j} k_4} + e^{q^T k_6}}
\end{aligned}
$}
\end{equation}

\subsubsection{Noise-aware Duplication}
While Eqn.\ref{eqn:goal2} suggests a straightforward solution, simply duplicating pruned keys and applying RoPE, it fails in practice, and will result in high-frequency visual artifacts. The root cause lies in the diffusion formulation: while the underlying content of two patches may be similar ($P_1 \approx P_2$), their corresponding inputs include independent Gaussian noise $\epsilon$. Consequently, directly duplicating a token replicates both the signal $P_1$ and its specific noise component $\epsilon_1$. This creates artificial correlation between tokens, violating the diffusion model's assumption of independent and identically distributed (i.i.d.) noise and causing noise amplification during attention calculation. To resolve this, we propose Noise-Aware Duplication. Our core strategy is to decouple the content from the noise: we replicate only the estimated clean content $P_1$, ensuring that the noise component remains non-repetitive. This preserves the video structure without violating the i.i.d. noise assumption.

We achieve this by duplicating the tokens using the closest (temporal) keys from the kv-cache. This is because all tokens in the kv-cache are clean due to an additional denosing step with zero noise level to form the cache. Note that when the duplication occurs in the kv-cache instead of the denosing block, we can still directly duplicate keys using the closest keys because it is already noise free. 

However, this produces a new problem; in the previous example, it is valid to have the approximatation $P_4 \approx P_5$. However, to avoid the interference of noise, we now use $P_1$ to approximate $P_5$, which does not hold. As a result, we put an extra constraint to the Latent Inter-Frame Pruning, such that a token is pruned if:
\begin{align}
\label{eqn:second_mask}
& \|P_{t^*}^{(x,y)} - P_{t}^{(x,y)}\|_1 < \theta_2, \\
& \text{where } t^* = 
    \begin{cases} 
        t - 1 & \text{if } t \equiv 0 \pmod 3 \\
        3 \lfloor t/3 \rfloor & \text{otherwise}
    \end{cases} 
\end{align}

This constraints enforces that the pruned token and its substituted token in the KV-cache must be similar.

\begin{figure}[htp]
  \centering
  \includegraphics[width=0.4\textwidth]{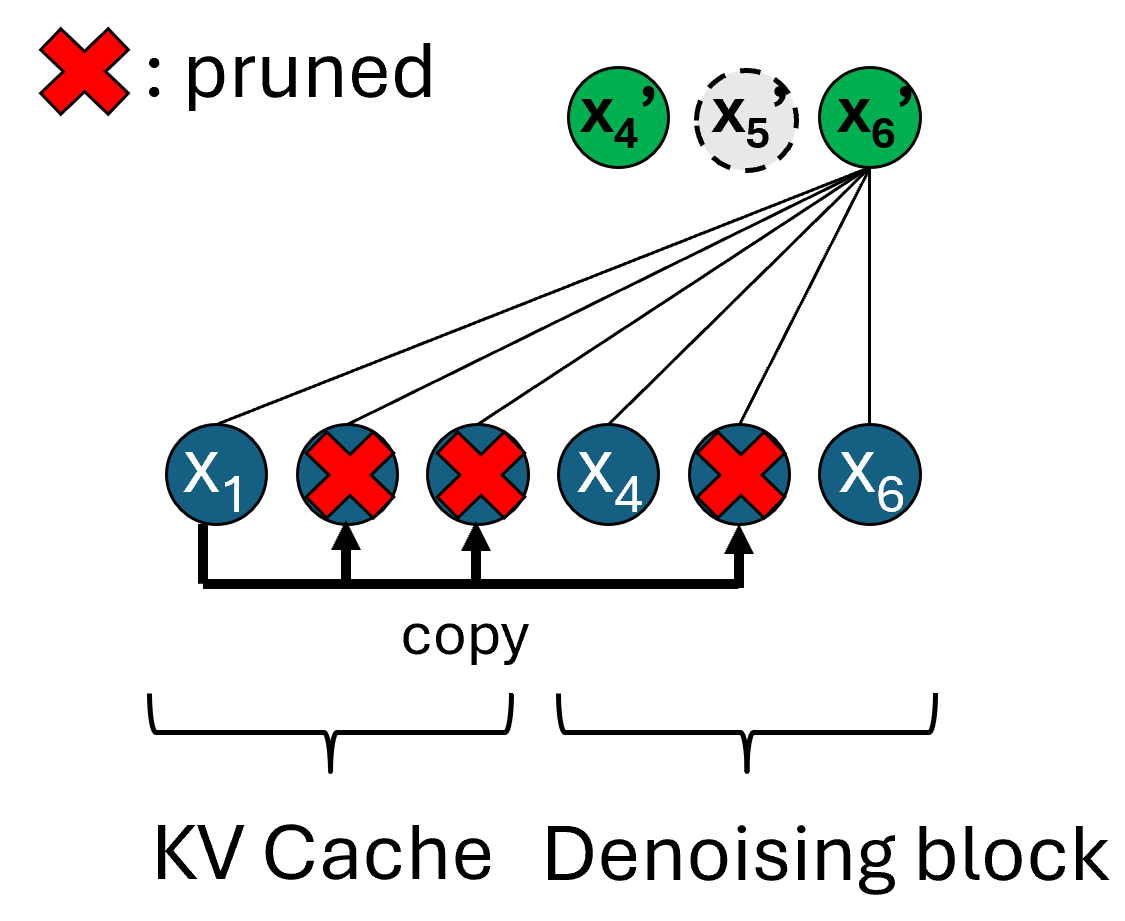}
  \caption{Noise-aware unpruning.e}
  \label{fig:noise_aware_unpruning2}
\end{figure}


\begin{table*}[!htb]
\centering
\caption{Quantitative comparison with the Baseline on the DAVIS dataset using VBench metrics. Results averaged over 51 edited videos demonstrate that our method maintains video quality comparable to the baseline.}
\label{tab:quality_comparison}
\begin{tabular}{lccccc}
\toprule
\textbf{Method} & \textbf{Subject} & \textbf{Background} & \textbf{Motion} & \textbf{Aesthetic} & \textbf{Imaging} \\
& \textbf{Consistency} & \textbf{Consistency} & \textbf{Smoothness} & \textbf{Quality} & \textbf{Quality} \\
\midrule
Baseline & 0.92 & 0.94 & 0.99 & 0.57 & 0.68 \\
Prune (Ours) & 0.92 & 0.94 & 0.99 & 0.58 & 0.67 \\
\bottomrule
\end{tabular}
\end{table*}

\subsubsection{Computational Acceleration}
In the following, we summarize how the video editing task is accelerated using our proposed pipeline:
\begin{enumerate}
\item \textbf{Generated Token Reduction:} Pruning reduces the total number of tokens needed to be generated (e.g., in Figure \ref{fig:math_goal} reducing generating $x_1$ to $x_6$ to only generate $\{x_1, x_4, x_6\}$. Reducing generating tokens directly accelerates all layers (Feed-Forward Networks, LayerNorm, Cross-Attention, and Self-Attention) in the Transformer and translates to a linear speedup proportional to the ratio $\frac{N_{total}}{N_{remain}}$ and is inherently compatible with hardware optimization libraries like FlashAttention and TensorRT due to the parallel processing of input tokens.
    \item \textbf{KV-Cache Compression:} Pruning redundant history tokens (e.g., removing $x_2$ from the cache) reduces the memory footprint and the computational cost of the attention mechanism itself. Since a smaller cache size reduces the complexity of the matrix multiplication, it further enhances generation speed.
\end{enumerate}

\subsubsection{Token Restoration}
After denoising, we will obtain pruned and cleaned tokens from the Diffusion Transformer. However,  to decode, the dimension for each frame is fixed. As a result, we must increase the number of tokens by duplicating clean tokens that contain similar content by referencing the prior frame. The implicit assumption here is that if the source video has similar patches, the edited video will also yield a similar denoised token. This holds especially the similarity is along the same spatial location as the neighboring temporal axis.

\section{Experiments}
We evaluated our pruning method using the Self-forcing model~\cite{huang2025selfforcing}, a fine-tuned variant of CausVid designed to bridge the training-inference gap and yield superior generation quality~\cite{yin2025causvid}. Experiments are conducted on the DAVIS dataset~\cite{davis2017dataset} using 51 video-text pairs, comparing our pruning approach against the baseline (full-sequence) model.  The pruning thresholds $\theta$ and $\theta_2$ (from Eq.~\ref{eq:pruning_criterion} and Eq.~\ref{eqn:second_mask}) were set to $0.15$ and $0.3$, respectively.

We assessed performance based on both video quality, using VBench, and inference throughput (FPS). Initial qualitative inspection shows that visual quality does not degrade despite pruning; to rigorously verify this, we compute VBench scores across all 51 edited videos. As shown in Table \ref{tab:quality_comparison}, the quantitative results confirm that the pruned model maintains quality comparable to the baseline. We then profiled inference throughput and found that the pruned method is, on average, $44.8\%$ faster than the baseline. The experimental results reinforce the effectiveness of our approach. 

\begin{figure}[htp]
  \centering
  \includegraphics[width=0.5\textwidth]{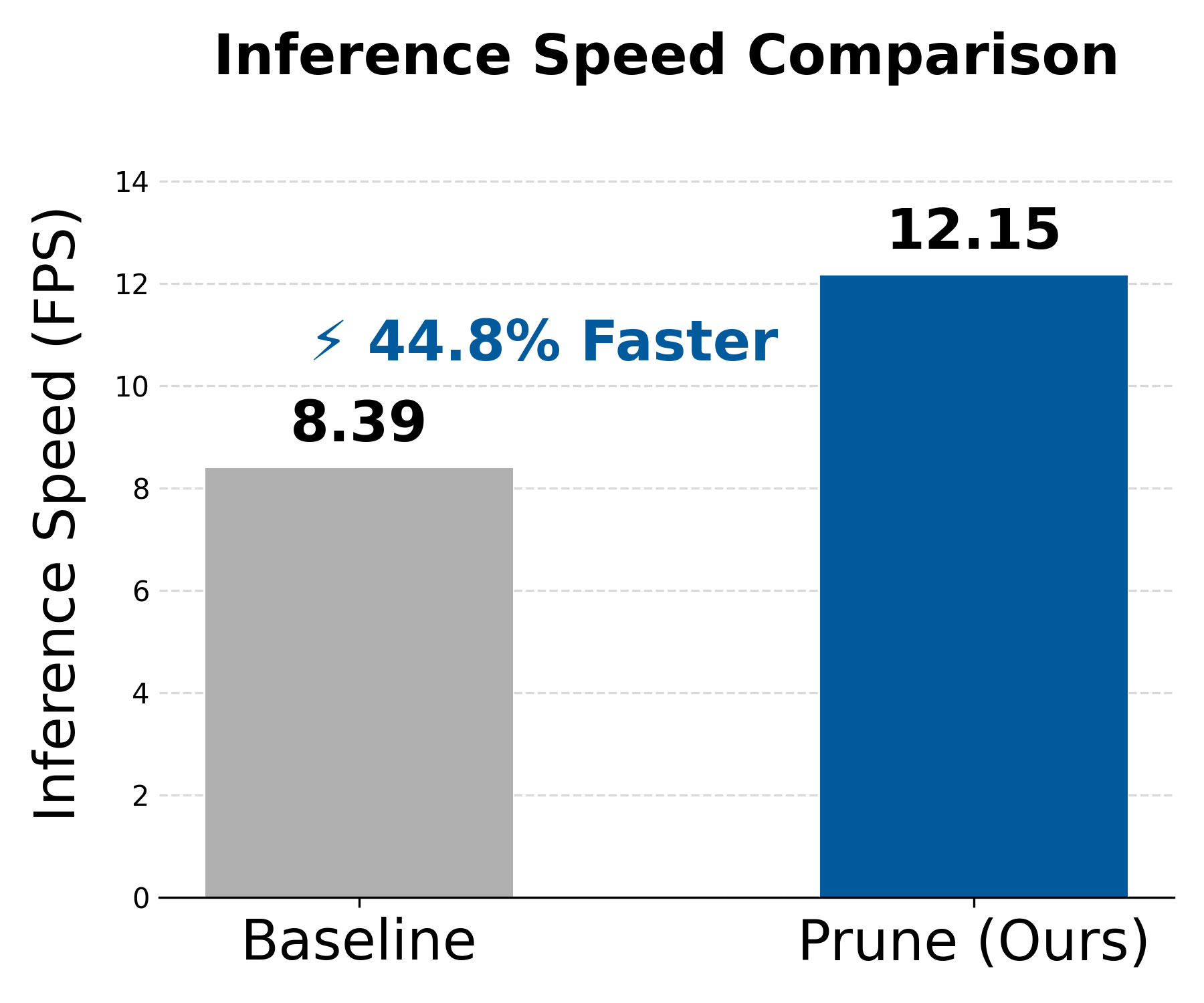}
  \caption{Inference speed comparisons between baseline and our pruning method.}
  \label{fig:noise_aware_unpruning}
\end{figure}





\section{Conclusion}
In this paper, we demonstrated the viability of integrating traditional video compression methods with modern video editing pipelines under the Latent Diffusion Model framework. Our method achieves a $1.44\times$ inference speedup while maintaining visual fidelity. Currently, our approach relies on fundamental Inter-Frame Pruning algorithm to bypass redundant computation. We believe that future work can achieve even higher efficiency by extending this paradigm to the spatial dimension and adopting adaptive strategies, such as variable pruning rates across different denoising steps. We hope this research inspires further exploration into end-to-end pruning for generative video models. 



{
    \small
    \bibliographystyle{ieeenat_fullname}
    \bibliography{main}
}

\end{document}